\documentclass[conference]{IEEEtran}
\IEEEoverridecommandlockouts

\usepackage{cite}
\usepackage{amsmath,amssymb,amsfonts}
\usepackage{algorithmic}
\usepackage{graphicx}
\usepackage{textcomp}
\usepackage{xcolor}
\usepackage{caption}
\usepackage{subcaption}
\usepackage{booktabs}
\usepackage{multirow}
\usepackage{makecell}
\usepackage{array}
\def\BibTeX{{\rm B\kern-.05em{\sc i\kern-.025em b}\kern-.08em
    T\kern-.1667em\lower.7ex\hbox{E}\kern-.125emX}}
\begin{document}

\title{\huge CAKE: Real-time Action Detection via Motion Distillation and Background-aware Contrastive Learning\\
}
\bstctlcite{IEEEexample:BSTcontrol}
\author{
  \IEEEauthorblockN{
    Hieu Hoang\IEEEauthorrefmark{2}\IEEEauthorrefmark{3},
    Dung Trung Tran\IEEEauthorrefmark{3},
    Hong Nguyen\IEEEauthorrefmark{4},
    Nam-Phong Nguyen\IEEEauthorrefmark{3}
  }
  \IEEEauthorblockA{
    \IEEEauthorrefmark{2}Viettel Group, Vietnam
  }
  \IEEEauthorblockA{
    \IEEEauthorrefmark{3}School of Electrical and Electronics Engineering, Hanoi University of Science and Technology, Vietnam
  }
  \IEEEauthorblockA{
    \IEEEauthorrefmark{4}Signal Analysis and Interpretation Lab, University of Southern California, United States
  }
}

\maketitle
\begin{abstract}
Online Action Detection (OAD) systems face two primary challenges: high computational cost and insufficient modeling of discriminative temporal dynamics against background motion. Adding optical flow could provides strong motion cues but it incurs significant computational overhead.
We propose CAKE, a OAD Flow-based distillation framework to transfer motion knowledge into RGB models. We propose Dynamic Motion Adapter (DMA) to suppress static background noise and emphasize pixel changes, effectively approximating optical flow without explicit computation. The framework also integrates a Floating Contrastive Learning strategy to distinguish informative motion dynamics from temporal background. Various experiments conducted on the TVSeries, THUMOS'14, Kinetics-400 datasets show effectiveness of our model. CAKE
achieves a standout mAP compared with SOTA while using the same backbone. Our model operates at over 72 FPS on a single CPU, making it highly suitable for resource-constrained systems.
\end{abstract}

\begin{IEEEkeywords}
Online Action Detection, Knowledge Distillation, Dynamic Convolution, Contrastive Learning, Edge Computing.
\end{IEEEkeywords}
\section{Introduction}
\label{sec:intro}

Online Action Detection (OAD) is important for real-time applications such as video surveillance, human–robot interaction, and autonomous driving. It aims to recognize ongoing actions from continuous video streams while using only past and current frames. Most existing methods rely on two assumptions: (i) motion must be obtained from optical flow, and (ii) all frames without an action can be grouped into a single ``background'' class. However, these assumptions do not fully match real video data because motion can often be inferred directly from RGB changes, and background frames usually contain many different visual states.

One of the key challenges in OAD is balancing motion modeling and computational efficiency, which is displayed in Tab. \ref{tab:runtime_comparison_fps}. Specifically, two-stream methods using RGB and optical flow often achieve strong performance by exploiting explicit motion signals \cite{xu2019trn,wang2021oadtr}. However, optical flow extraction requires excessive computation and inference latency, which limits real-time deployment. Moreover, RGB-only models are much faster but usually lack clear motion representations \cite{simonyan2014two}. Recently, knowledge distillation, where an RGB model learns motion features from an optical-flow teacher \cite{d3d2021}, has been regarded as a promising alternative. However, most existing approaches still rely on static convolutions, which cannot adapt well to the diverse motion patterns in videos \cite{huang2022tada}.

Moreover, background modeling introduces another difficulty, which is background diversity. In a video, action frames form only a small portion of the timeline, while the remaining frames are typically labeled as ``background''. However, these background frames do not represent a single semantic category. Instead, they include different situations such as action transitions, unrelated movements, and varied scene contexts. Treating them as one class forces many methods, including supervised contrastive learning (SupCon) \cite{khosla2020supcon}, to group very different samples, which can degrade the model's discriminative power.
These observations suggest that background frames should not be forced into a single feature cluster. Instead, they should be allowed to spread naturally in the feature space to reflect the diversity of video content. At the same time, extracting motion cues from RGB requires more flexible operations than static convolutions.

To address these challenges, we propose CAKE, a lightweight framework for online action detection that relies only on RGB inputs during inference. Our method consists of two main components. First, we design the Dynamic Motion Adapter (DMA), which generates adaptive convolutional weights to emphasize pixel variations while suppressing responses to static regions. This mechanism allows the network to capture motion cues without explicitly computing optical flow. Second, we adopt a contrastive learning strategy that applies clustering constraints only to action samples, while background frames are allowed to distribute naturally.

Our contributions are summarized as follows:
\begin{itemize}
    \item A DMA based on omni-dimensional spatio-temporal dynamic convolution (ODConv3D) is proposed. Through multi-modal knowledge distillation, the network learns motion-aware representations directly from RGB inputs without explicit optical flow inference.

    \item Floating SupCon is introduced, where action classes are clustered while background frames remain unconstrained.

    \item Extensive experiments show the effectiveness and scalability of the proposed framework through two variants, including a high-accuracy two-stream model (CAKE-R50) and an efficient RGB-only model (CAKE-X3D).
\end{itemize}
\section{Related Work}
\label{sec:related_work}

Early research on Online Action Detection (OAD) mainly relied on recurrent architectures. Temporal Recurrent Network (TRN)~\cite{xu2019trn} leverages recurrent structures to accumulate long-term temporal context for frame-level prediction in streaming videos. IDN~\cite{eun2020idn} improves this framework by introducing an information discrimination mechanism that selectively filters useful historical cues while suppressing irrelevant signals.

To better model long-range temporal dependencies, recent methods adopt Transformer-based architectures. OadTR~\cite{wang2021oadtr} is among the first to introduce self-attention for OAD, enabling global temporal reasoning over past observations. LSTR~\cite{xu2021lstr} extends this idea by separating video history into long-term and short-term memories, allowing efficient modeling of long video sequences. GateHUB~\cite{chen2022gatehub} further enhances historical encoding through a gated mechanism that emphasizes informative frames while suppressing noisy background signals. Despite their strong performance, transformer-based approaches often incur high computational costs.
To improve efficiency, MiniROAD~\cite{miniroad2023} proposes a minimalist RNN framework with uneven loss weighting, reducing the discrepancy between training and inference while maintaining competitive performance with significantly lower computational cost. However, achieving high accuracy often still relies on two-stream architectures that require explicit optical flow extraction.

Another line of work explores multimodal knowledge distillation. Methods such as D3D~\cite{d3d2021} transfer motion knowledge from an optical-flow-based teacher network to an RGB-only student model, enabling the use of motion cues without computing optical flow during inference. Nevertheless, many existing motion modeling modules rely on static 3D convolutions with fixed filters, limiting their ability to adapt to diverse motion patterns in real-world videos.
Moreover, SupCon has also been explored to improve representation learning. However, in the OAD setting, the severe imbalance between action frames and background frames makes it difficult to enforce a single compact cluster for all background samples without degrading the discriminative structure of the feature space.

Unlike the approaches above, our method focuses on learning flexible motion representations directly from RGB inputs via a DMA. In addition, we introduce Floating SupCon that allows background frames to distribute naturally in the feature space rather than forcing them into a single cluster.
\section{Methodology}
\label{sec:Method}
\begin{figure*}
    \includegraphics[width=1\textwidth]{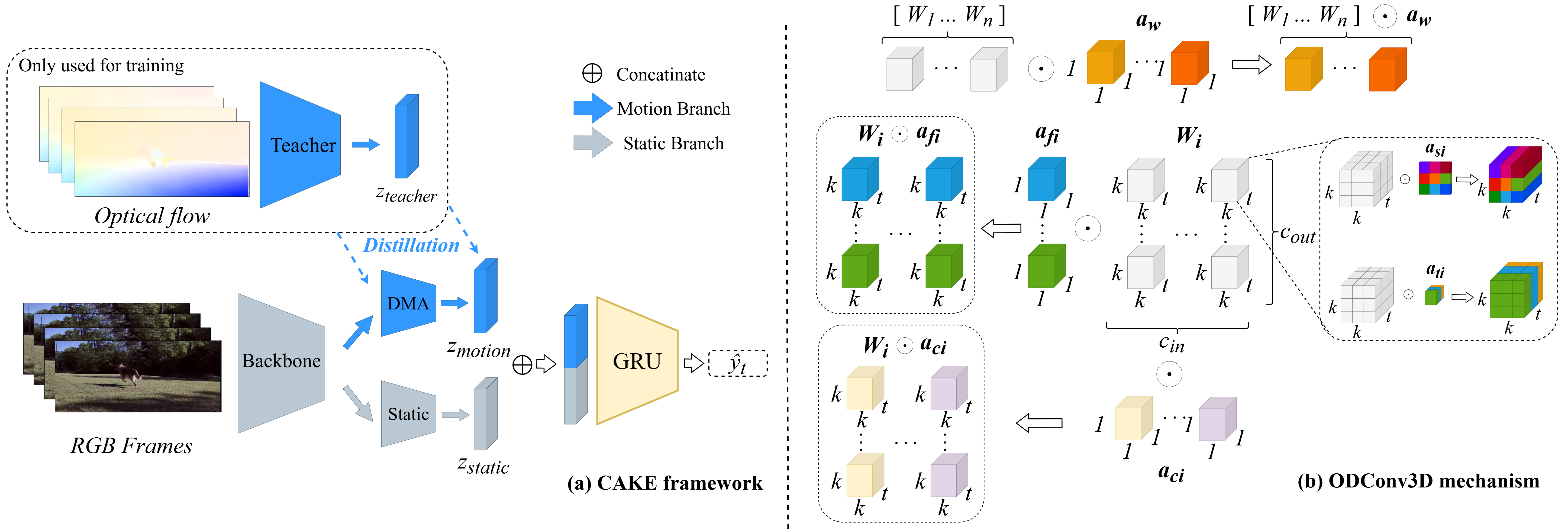}
    \caption{Overview of the proposed CAKE framework and DMA. \textbf{(a)} During training, an optical-flow teacher guides the DMA via cross-modal distillation. At inference, the RGB backbone splits into a static branch and a DMA motion branch, whose features are fused before temporal modeling with a GRU. \textbf{(b)} ODConv3D mechanism used in DMA, where base kernels $W_i$ are modulated by dynamic attention weights along kernel ($a_w$), channel ($a_f, a_c$), spatial ($a_s$), and temporal ($a_t$) dimensions.}
    \label{fig:architecture}
\end{figure*}
Our goal is to recognize actions at the current time step using only past observations.
Theoretically, the task is to use the sequence up to $v_t$ to predict the probability distribution $\hat{y}_t \in \{0, 1, \dots, K\}$, where $K$ denotes the number of action classes and $0$ represents the background class. During training, each video is divided into clips of length $L$, and the segment from $t-L+1$ to $t$ is used as input. Following standard practice, CAKE adopts a bifurcated spatial backbone to encode each frame $v_t$ into a multi-modal feature vector $z_t \in \mathbb{R}^C$. This feature at time $t$ is then used to model temporal dependencies and produce the final prediction $\hat{y}_t$.

\subsection{Motion Distillation via Bifurcated Backbone}
Traditional architectures face a fundamental trade-off: two-stream networks that combine RGB and optical flow achieve strong accuracy but incur substantial computational cost, while single-stream RGB models lack explicit motion cues. To address this challenge, CAKE introduces a bifurcated feature extractor that encodes motion information directly from RGB frames through cross-modal distillation.

\noindent
\textbf{Bifurcated Student Architecture.} Let $v_t$ denote the input RGB frame at time $t$ and $F_{backbone}$ represent the frozen convolutional layers of the spatial backbone. We extract an intermediate representation through $z_{static} = F_{backbone}(v_t)$. This feature purely represents spatial context. To model motion, a parallel auxiliary branch is constructed using the DMA by $z_{motion} = DMA(z_{static})$.
These two information streams are subsequently fused to form the multi-modal representation at each time step by concatinating $z_{static}$ and $z_{motion}$.
This design enables the network to simultaneously learn static spatial and temporal patterns without mutual interference.

\vspace{1ex}
\noindent
\textbf{DMA Module with ODConv3D.} The motion branch is built upon an ODConv3D, which extends the original ODConv operator \cite{li2022odconv} to the spatio-temporal domain. Unlike conventional convolutions that use fixed filters, ODConv3D generates adaptive convolutional weights conditioned on the static feature $z_{static}$. 
Assume $n$ base convolution kernels $W_i$ ($i \in \{1, \dots, n\}$). The dynamic weight for each kernel is
$W_{dyn}^{(i)} = \alpha_{wi} \odot \alpha_{fi} \odot \alpha_{ci} \odot \alpha_{si} \odot \alpha_{ti} \odot W_i$,
where the attention scalars control the importance of the kernel ($w$), input channel ($f$), output channel ($c$), spatial ($s$), and temporal ($t$) dimensions.
By adjusting these factors jointly, the convolution filter can adapt to different motion patterns.

\noindent
\textbf{Distillation from Optical Flow.}~To effectively learn the motion information with out optical flow itself, we apply a knowledge distillation mechanism. During training, we train a computationally heavy Teacher model with actual Optical Flow data ($X_{flow}$) to extract standard motion features:
\begin{equation}
z_{Teacher} = Teacher(X_{flow}),
\end{equation}
The DMA branch of the Student Network is then optimized to approximate this representation space via a distillation loss:
\begin{equation}
\mathcal{L}_{distill} = ||z_{Teacher} - z_{motion}||_2^2.
\end{equation}
The synergy between ODConv3D and this structural loss completely isolates the computationally expensive motion-simulation step from the inference phase.

\subsection{Training-Inference Discrepancy}
Once extracted, the feature sequence $z_t$ is fed into the recurrent module (GRU).
During online inference, the GRU progressively accumulates temporal information while during training, videos are truncated into chunks of length $L$, which disconnects the initial hidden state from prior context.
CAKE integrates the temporal alignment strategy from \cite{miniroad2023}. We employ a step function to mask the loss weight ($\alpha_t$) so that it activates exclusively at the final time step:
\begin{equation}
\alpha_t = \begin{cases} 1, & \text{if } t = L \\ 0, & \text{otherwise} \end{cases}, \quad 
\mathcal{L}_{temporal} = \sum_{t=1}^{L} \alpha_t \ell(\hat{y}_t, y_t)
\end{equation}
This mechanism discards unreliable early predictions and compels the GRU to learn only from hidden states that have accumulated sufficient temporal context over the sequence length $L$, thereby better aligning training with inference conditions.

\subsection{Floating Contrastive Learning}

To refine the representation space from the GRU output sequence, we introduce Floating SupCon, built upon the Momentum Contrast framework \cite{he2020moco}.
The GRU outputs are projected into query vectors ($q$). Their momentum copies pass through a momentum encoder to generate positive keys ($k_+$) and populate a dictionary queue $\mathcal{Q}$. The momentum encoder is updated using Exponential Moving Average (EMA) from the query network, ensuring key consistency across mini-batches.

Standard SupCon \cite{khosla2020supcon} enforces homogeneous clustering across all classes. However, in OAD the background class is highly heterogeneous, and forcing all background frames into a single cluster can distort the representation space. To address this issue, Floating SupCon introduces an asymmetric contrastive objective.

\begin{figure}[htbp]
    \centering
    \includegraphics[width=0.8\linewidth]{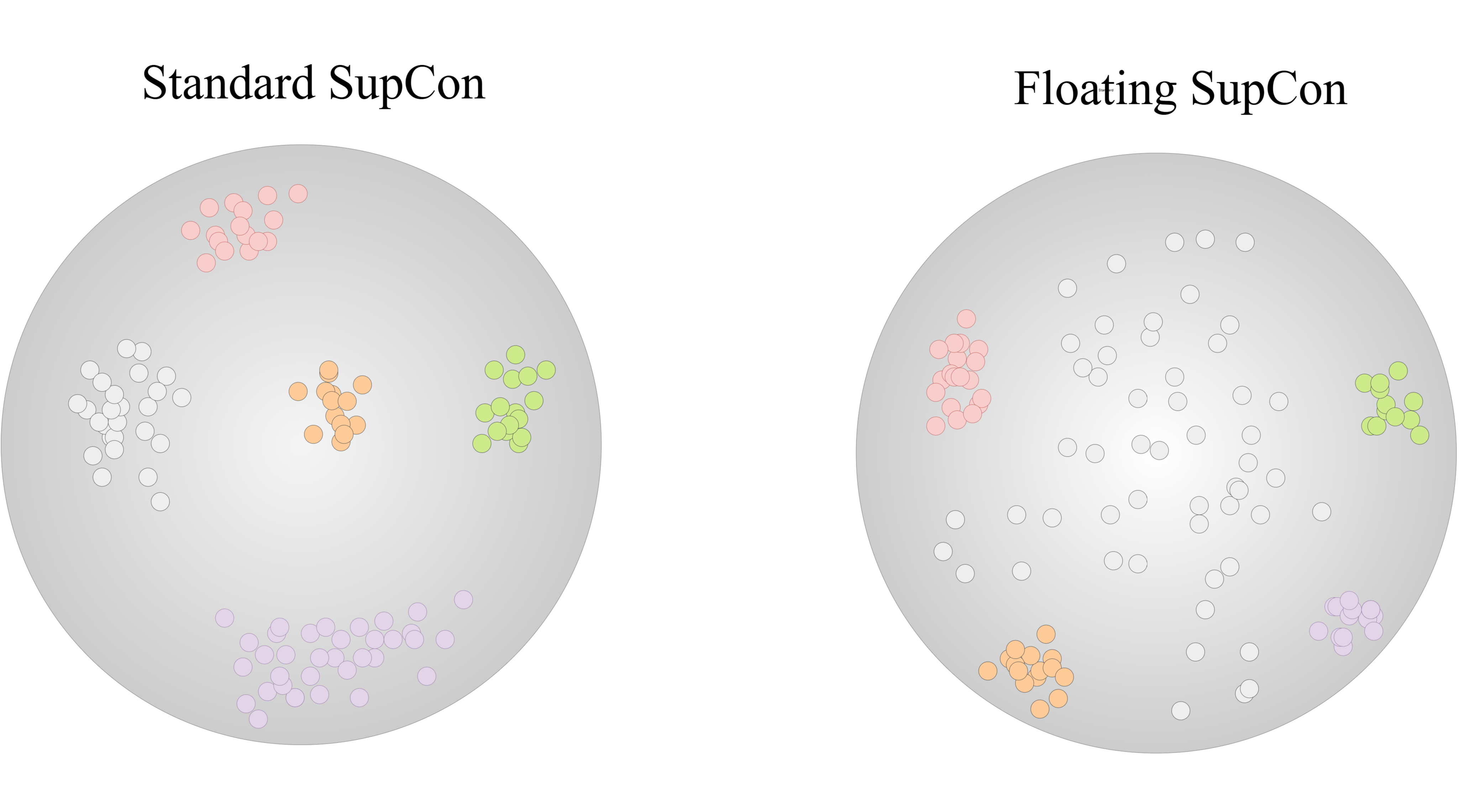}
    \caption{Overview of the MoCo mechanism. The query encoder is updated by backpropagation, while the momentum encoder is updated using Exponential Moving Average (EMA) to maintain a consistent dictionary queue.}
    \label{fig:moco_framework}
\end{figure}

\noindent\textbf{Action Clustering.} 
For action queries ($y_q \neq y_b$), the model pulls the query ($q$) toward samples of the same action class and pushes it away from other samples:

\begin{equation}
\mathcal{L}_{act}(q) =
- \log
\frac{S(q,k_+) + \sum_{k \in \mathcal{Q}: y_k = y_q} S(q,k)}
{S(q,k_+) + \sum_{k' \in \mathcal{Q}} S(q,k')},
\end{equation}

where $y_b$ denotes the background label, $k_+$ denotes the momentum copy of $q$, $k$ and $k'$ are keys stored in the queue $\mathcal{Q}$, and $S(a,b)=\exp(a \cdot b/\tau)$ denotes the temperature-scaled cosine similarity.

\noindent\textbf{Background Floating.} 
For background queries ($y_q = y_b$), the query is only aligned with its own momentum copy ($k_+$) and repelled from action samples. Importantly, background samples are not forced to cluster with each other:

\begin{equation}
\mathcal{L}_{bg}(q) =
- \log
\frac{S(q,k_+)}
{S(q,k_+) + \sum_{k' \in \mathcal{Q}: y_{k'} \neq y_b} S(q,k')}.
\end{equation}

This design allows background representations to spread naturally in the feature space while preserving action vectors.

\section{Experiments and Results}
\label{sec:experiments_results}

\subsection{Datasets and Implementation Details}

\noindent
\textbf{Datasets.} 
We evaluate CAKE on two public online action detection benchmarks: THUMOS'14 \cite{idrees2017thumos}, which contains unconstrained sports videos with 20 action classes, and TVSeries \cite{degeest2016online}, which consists of 16 hours of real-life videos with 30 actions and complex background transitions. In addition, we pre-train the Dynamic Motion Adapter (DMA) on Kinetics-400 \cite{carreira2017i3d} to learn generalized motion representations.

\vspace{1ex}
\noindent
\textbf{Model Settings.}
We adopt X3D \cite{feichtenhofer2020x3d} as the backbone, which processes 13-frame clips and outputs 192-dimensional spatial feature vectors. The DMA employs depthwise separable convolutions, consisting of a $3\times3\times3$ depthwise filter followed by a $1\times1\times1$ pointwise filter. The flow hallucination block further factorizes the operation into temporal ($3\times1\times1$) and spatial ($1\times3\times3$) filters. The reduction ratio for the attention module in all ODConv3D layers is set to $1/16$.

The resulting feature sequence is fed into a single-layer GRU with a hidden size of 1024. The GRU training strategy follows MiniROAD \cite{miniroad2023}, while the Floating SupCon module adopts the hyperparameters from the MoCo \cite{he2020moco} framework.

Training is performed in three stages: 
(1) pre-training the backbone and DMA on Kinetics-400 for 100 epochs using SGD with an initial learning rate of 0.1; 
(2) freezing the backbone and training the GRU on the target datasets for 50 epochs using AdamW; 
and (3) fine-tuning the final linear classifier.

To address the severe class imbalance between action and background frames, the final stage replaces standard cross-entropy with Focal Loss \cite{lin2017focal}:
\begin{equation}
\mathcal{L}_{focal} = - \alpha_t (1 - p_t)^\gamma \log(p_t)
\end{equation}

The model is implemented in PyTorch and trained on a single Nvidia RTX 3090 GPU.

\vspace{1ex}
\noindent
\textbf{Evaluation Metrics.}~We report per-frame mean Average Precision (mAP) on THUMOS'14. For TVSeries, we use mean calibrated Average Precision (mcAP) \cite{degeest2016online}.

\subsection{Comparison with State-of-the-Art}

\begin{table}[htbp]
\centering
\caption{Performance and efficiency comparison of CAKE and SOTA methods on THUMOS'14 and TVSeries. \textit{BNI}: BN-Inception.}
\label{tab:comprehensive_comparison}
\setlength{\tabcolsep}{2.5pt}
\resizebox{\columnwidth}{!}{%
\begin{tabular}{@{} l l c c c c c @{}}
\toprule
\multirow{2}{*}{\textbf{Method}} & \multirow{2}{*}{\textbf{Backbone}} & \multicolumn{2}{c}{\textbf{Params (M)}} & \multirow{2}{*}{\textbf{GFLOPs}} & \textbf{THUMOS'14} & \textbf{TVSeries} \\
\cmidrule(lr){3-4}
 & & \textbf{Back.} & \textbf{Head} & & \textbf{mAP (\%)} & \textbf{mcAP (\%)} \\
\midrule
RED \cite{gao2017red} & - & - & - & - & 45.3 & 79.2 \\
LAP \cite{qu2020lap} & - & - & - & - & 53.3 & 85.3 \\
TFN \cite{eun2021tfn} & - & - & - & - & 55.7 & 85.0 \\
FATS \cite{kim2021fats} & - & - & - & - & 59.0 & 84.6 \\
IDN \cite{eun2020idn} & - & - & - & - & 60.3 & 86.1 \\
TRN \cite{xu2019trn} & R200+BNI & 72.9 & 402.9 & 104.6 & 62.1 & 86.2 \\
PKD \cite{zhao2020pkd} & - & - & - & - & 64.5 & 86.4 \\
OadTR \cite{wang2021oadtr} & R200+BNI & 72.9 & 75.8 & 105.6 & 65.2 & 87.2 \\
Colar \cite{yang2022colar} & - & - & - & - & 66.9 & 88.1 \\
LSTR \cite{xu2021lstr} & R50+BNI & 33.8 & 58.0 & 44.6 & 69.5 & 89.1 \\
GateHUB \cite{chen2022gatehub} & R50+BNI & 33.8 & 45.2 & 44.1 & 70.7 & 89.6 \\
TeSTra \cite{zhao2022testra} & R50+BNI & 33.8 & 58.9 & 41.4 & 71.2 & - \\
MiniROAD \cite{miniroad2023} & R50+BNI & 33.8 & 15.8 & 37.1 & 71.4 & 89.6 \\
MALT \cite{yang2024malt} & R50+BNI & 33.8 & - & 37.1 & 71.4 & \textbf{89.7} \\
\midrule
CAKE-R50 (Ours) & R50+BNI & 33.8 & 15.8 & 37.1 & \textbf{72.0} & - \\
CAKE-X3D (Ours) & X3D (RGB) & \textbf{2.9} & 17.5 & \textbf{2.9} & 67.1 & 86.5 \\
\bottomrule
\end{tabular}%
}
\end{table}

\noindent
\textbf{Comparison with State-of-the-Art.}
Table \ref{tab:comprehensive_comparison} summarizes both performance and efficiency comparisons with prior OAD methods. Under the same backbone configuration (R50+BNI), CAKE achieves the best performance on THUMOS'14 while maintaining comparable model complexity to recent baselines. This improvement indicates that the proposed Floating SupCon strategy successfully organizes the representation space by preventing heterogeneous background frames from collapsing into a single cluster. {These results demonstrate that our contrastive formulation improves action--background separability without increasing computational overhead.

Furthermore, the lightweight CAKE-X3D achieves competitive accuracy using only RGB inputs despite having substantially fewer parameters and lower computational complexity than conventional two-stream models. This result suggests that the Dynamic Motion Adapter (DMA) can effectively capture essential motion cues directly from RGB features, providing a practical alternative to explicit optical flow extraction. This observation highlights the representation efficiency of the proposed dynamic motion modeling mechanism.

When evaluated on the more challenging TVSeries dataset, CAKE maintains competitive performance under complex real-world conditions with frequent background transitions. The consistent results across datasets indicate that the proposed framework generalizes well to diverse video environments. Overall, these results confirm that CAKE achieves a favorable balance between accuracy, efficiency, and robustness for online action detection.

\begin{figure*}[htbp]
    \centering
    \includegraphics[width=0.24\textwidth]{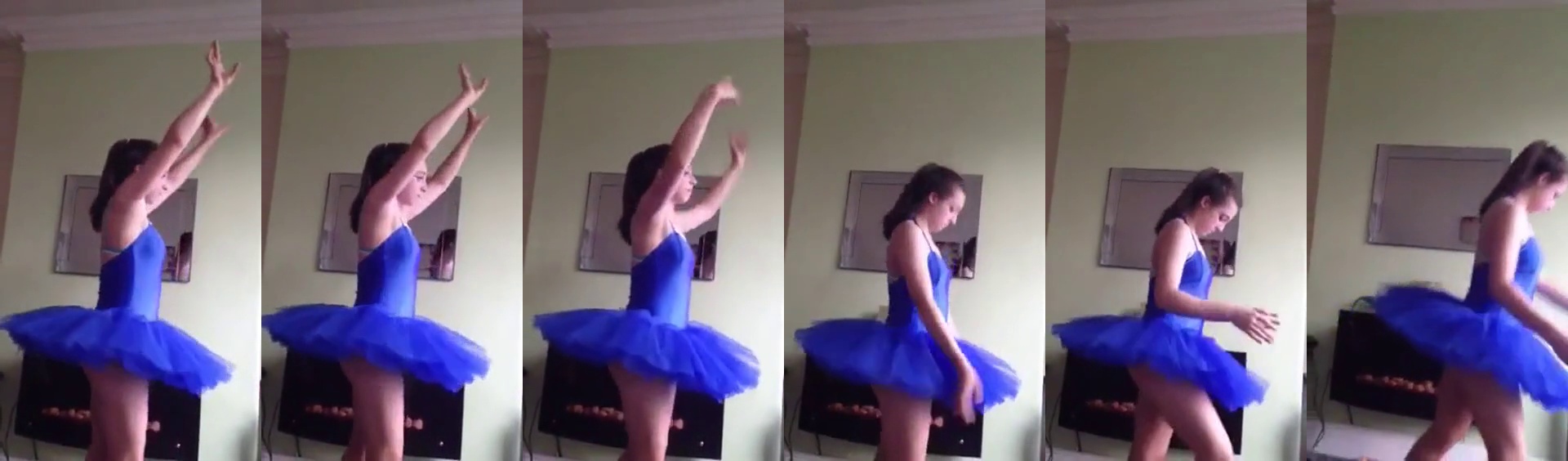}\hfill
    \includegraphics[width=0.24\textwidth]{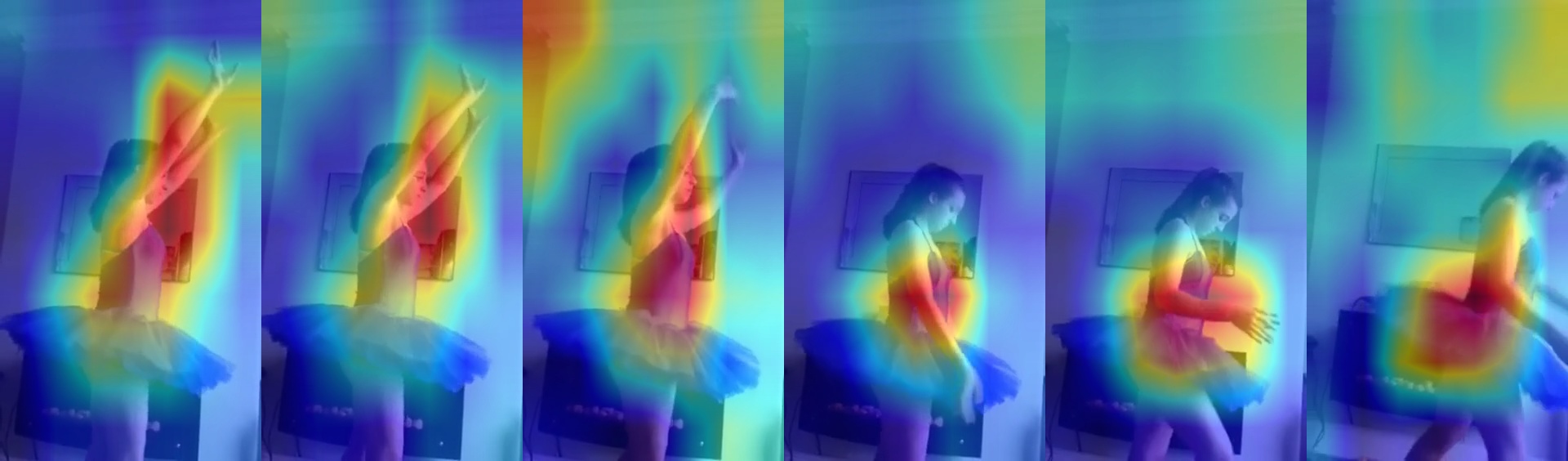}\hfill
    \includegraphics[width=0.24\textwidth]{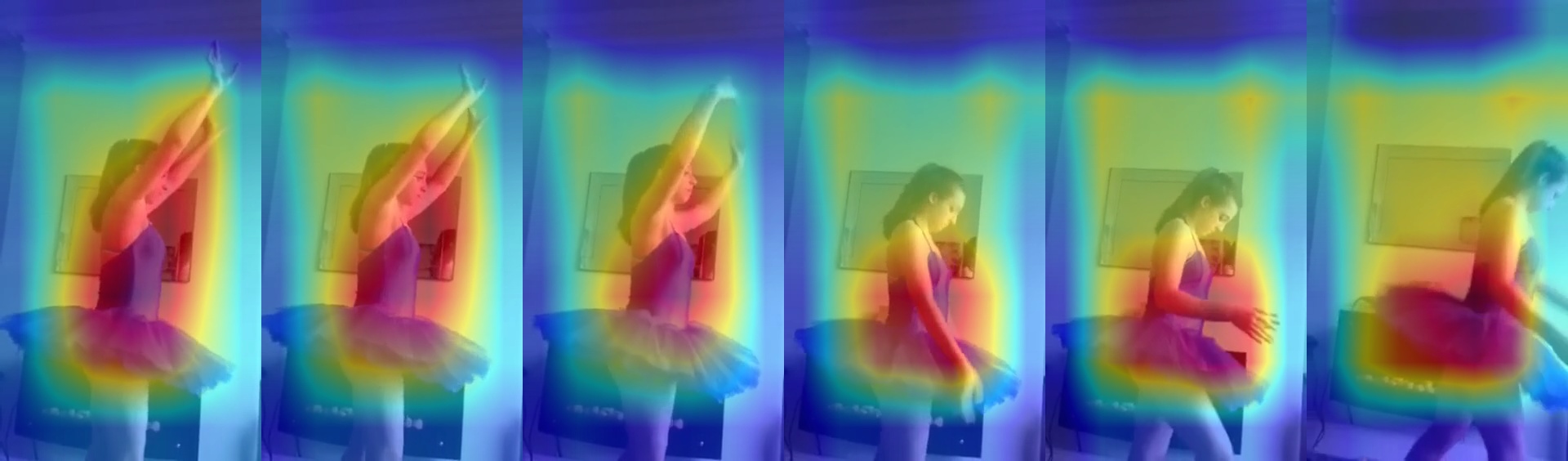}\hfill
    \includegraphics[width=0.24\textwidth]{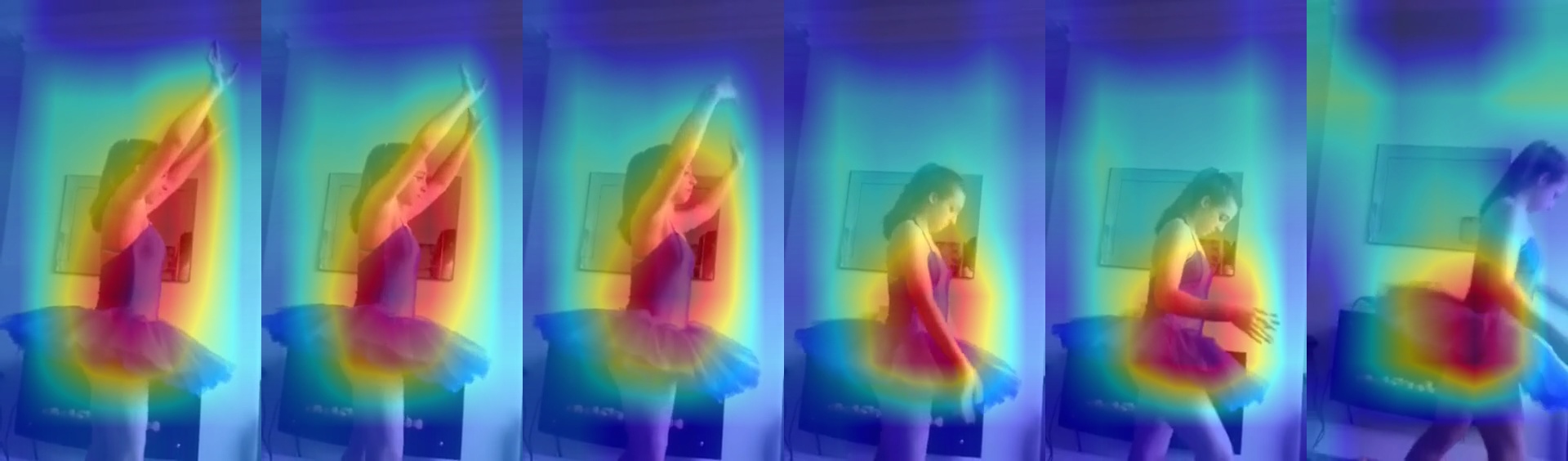}
    \caption{Qualitative analysis using Grad-CAM. From left to right: (a) Original RGB frames, (b) Teacher trained on Optical Flow, (c) RGB Student with static Conv3D, and (d) Our DMA with ODConv3D.}
    \label{fig:gradcam_analysis}
\end{figure*}

\begin{table}[t]
\centering
\caption{Detailed inference speed (FPS) analysis. `$\times$' denotes non-real-time execution. OF: Optical Flow.}
\label{tab:runtime_comparison_fps}
\resizebox{\linewidth}{!}{%
\begin{tabular}{lccccccc}
\toprule
\textbf{Method} & 
\textbf{\makecell{OF \\ Comp.}} & 
\textbf{\makecell{RGB \\ Feat.}} & 
\textbf{\makecell{OF \\ Feat.}} & 
\textbf{\makecell{Model \\ Head}} & 
\textbf{\makecell{Overall \\ FPS \\ (GPU)}} & 
\textbf{\makecell{Overall \\ FPS \\ (CPU)}} & 
\textbf{\makecell{mAP \\ (\%)}} \\ 
\midrule
LSTR \cite{xu2021lstr}       & \multirow{6}{*}{28.8} & \multirow{6}{*}{383} & \multirow{6}{*}{219} & 187   & 21.2 & $\times$ & 69.5 \\
TRN \cite{xu2019trn}        &                       &                      &                      & 143   & 20.5 & $\times$ & 62.1 \\
OadTR \cite{wang2021oadtr}    &                       &                      &                      & 145   & 20.5 & $\times$ & 65.2 \\
TeSTra \cite{zhao2022testra} &                       &                      &                      & 169   & 20.9 & $\times$ & 71.2 \\
MiniROAD \cite{miniroad2023} &                       &                      &                      & \textbf{37,300} & 23.8 & $\times$ & 71.4 \\
\midrule
CAKE-R50 (Ours)      &                       &                      &                      & \textbf{37,300} & 23.8 & $\times$ & \textbf{72.0} \\
CAKE-X3D (Ours)      & -                     & \textbf{515}         & -                    & \textbf{37,300} & \textbf{$>$100} & \textbf{72.8} & 67.1 \\
\bottomrule
\end{tabular}%
}
\end{table}

\vspace{1ex}
\noindent
\textbf{Runtime Analysis.}~Tab. \ref{tab:runtime_comparison_fps} highlights a key limitation of traditional two-stream frameworks: the expensive TV-L1 \cite{zach2007duality} optical flow extraction stage significantly constrains the overall inference speed, preventing real-time deployment on standard CPUs. By eliminating this preprocessing step, CAKE-X3D enables a streamlined inference pipeline while maintaining competitive detection accuracy. This efficiency gain makes the proposed framework more suitable for practical real-time deployment on resource-constrained devices.

\subsection{Ablation Study}

We conduct ablation studies to quantify the individual contributions of the proposed components. Table \ref{tab:component_ablation} reports the performance changes when progressively integrating the Dynamic Motion Adapter (DMA) and the contrastive learning module into the basic RGB pipeline.

\begin{table}[htbp]
\centering
\caption{Ablation study of individual components.}
\label{tab:component_ablation}
\resizebox{\columnwidth}{!}{%
\begin{tabular}{ccccc}
\toprule
\textbf{RGB (Static)} & \textbf{Flow (DMA)} & \textbf{Contrastive} & \textbf{THUMOS'14} & \textbf{TVSeries} \\
& & & \textbf{mAP (\%)} & \textbf{mcAP (\%)} \\
\midrule
\checkmark & & & 63.1 & 80.21 \\
\checkmark & \checkmark & & 66.0 \textcolor{blue}{(+2.9)} & 85.72 \textcolor{blue}{(+5.51)} \\
\checkmark & \checkmark & \checkmark & \textbf{67.1 \textcolor{blue}{(+1.1)}} & \textbf{86.50 \textcolor{blue}{(+0.78)}} \\
\bottomrule
\end{tabular}%
}
\end{table}

\vspace{1ex}
\noindent
\textbf{Effectiveness of Motion Distillation.}
Using only static RGB features yields limited performance, particularly on the TVSeries dataset where complex background dynamics frequently occur. Introducing the DMA module consistently improves detection accuracy across both datasets, suggesting that motion-aware features are critical for reliable online action recognition. By dynamically adapting convolutional filters, DMA enables the network to focus on temporally relevant regions rather than static background patterns. This result confirms that motion distillation substantially improves motion sensitivity within a single RGB stream.

\vspace{1ex}
\noindent
\textbf{Impact of Floating SupCon.}
The contrastive learning module further enhances performance by restructuring the learned representation space. As shown in Table \ref{tab:supcon_comparison}, applying the standard supervised contrastive loss \cite{khosla2020supcon} leads to a slight performance degradation relative to the baseline. This occurs because conventional SupCon forces highly diverse background frames into a single compact cluster, which distorts the feature distribution. In contrast, the proposed Floating SupCon clusters action classes while allowing background samples to remain unconstrained. This produces a more semantically structured feature space and achieves good performance.

\begin{table}[htbp]
\centering
\caption{Performance comparison between MiniROAD, Standard SupCon, and the proposed Floating SupCon on THUMOS'14.}
\label{tab:supcon_comparison}
\resizebox{\columnwidth}{!}{%
\begin{tabular}{llc}
\toprule
\textbf{Method} & \textbf{Learning Mechanism} & \textbf{mAP (\%)} \\
\midrule
MiniROAD \cite{miniroad2023} & Supervised (RNN) & 71.40 \\
CAKE (Baseline) & Standard SupCon & 70.80 \textcolor{red}{(-0.6)} \\
CAKE (Ours) & Floating SupCon & \textbf{72.00 \textcolor{blue}{(+0.6)}} \\
\bottomrule
\end{tabular}%
}
\end{table}

\subsection{Motion Representation Learning Evaluation}
\label{sec:motion_representation}
This section investigates whether an RGB-based model can learn motion representations comparable to those of networks explicitly trained on optical flow. The evaluation is performed on the Kinetics-400 benchmark, with results summarized in Table \ref{tab:kinetics_ablation}.
\begin{table}[htbp]
\centering
\caption{Performance comparison on Kinetics-400. `*' denotes optical flow reconstruction accuracy. R: RGB, OF: Optical Flow.}
\label{tab:kinetics_ablation}
\setlength{\tabcolsep}{3pt} 
\resizebox{\columnwidth}{!}{%
\begin{tabular}{@{} l l c c c c c c @{}} 
\toprule
\multirow{2}{*}{\textbf{Model}} & \multirow{2}{*}{\textbf{Input}} & \multicolumn{3}{c}{\textbf{Top-1 Acc. (\%)}} & \multirow{2}{*}{\textbf{Top-5}} & \multirow{2}{*}{\textbf{\makecell{GFLOPs \\ $\times$ views}}} & \multirow{2}{*}{\textbf{\makecell{Params \\ (M)}}} \\
\cmidrule(lr){3-5}
 & & \textbf{RGB} & \textbf{Flow} & \textbf{Comb.} & & & \\
\midrule
\multicolumn{8}{@{}l}{\textit{Single-stream Models}} \\
I3D \cite{carreira2017i3d} & R & 71.10 & - & - & 90.30 & $108.0 \times \mathrm{N/A}$ & 12.0 \\
C2D R50 \cite{wang2018nonlocal} & R & 71.46 & - & - & 89.68 & $25.89 \times 30$ & 24.3 \\
Slow R50 ($4\times16$) \cite{feichtenhofer2019slowfast} & R & 72.40 & - & - & 90.18 & $27.55 \times 30$ & 32.4 \\
X3D-S (Baseline) \cite{feichtenhofer2020x3d} & R & 73.33 & - & - & 91.27 & $1.96 \times 30$ & 3.8 \\
TSM R50 \cite{lin2019tsm} & R & 74.70 & - & - & N/A & $65.0 \times 10$ & 24.3 \\
X3D-L \cite{feichtenhofer2020x3d} & R & 77.44 & - & - & 93.31 & $26.64 \times 30$ & 6.2 \\
MViT-B ($16\times4$) \cite{fan2021mvit} & R & 78.85 & - & - & 93.85 & $70.80 \times 5$ & 36.6 \\
\midrule
\multicolumn{8}{@{}l}{\textit{Two-stream Models}} \\
SlowFast R50 \cite{feichtenhofer2019slowfast} & R+OF & - & - & 75.34 & 91.89 & $36.69 \times 30$ & 34.5 \\
Two-Stream I3D \cite{carreira2017i3d} & R+OF & - & - & 75.70 & 92.00 & $216.0 \times \mathrm{N/A}$ & 25.0 \\
Two-Stream S3D-G \cite{xie2018rethinking} & R+OF & - & - & 77.20 & 93.00 & $143.0 \times \mathrm{N/A}$ & 23.1 \\
\midrule
\multicolumn{8}{@{}l}{\textit{CAKE Variants (Ours)}} \\
Teacher X3D-S & OF & - & 50.21 & - & - & $1.96 \times 30$ & 3.8 \\
CAKE (Conv3D) & R & 73.33 & 43.21* & - & 91.27 & $2.85 \times 30$ & 6.1 \\
CAKE (ODConv3D) & R & \textbf{73.33} & \textbf{44.71*} & \textbf{74.20} & \textbf{92.30} & \textbf{$2.96 \times 30$} & 6.4 \\
\bottomrule
\end{tabular}%
}
\end{table}
To objectively evaluate motion learning, we directly assess the DMA branch using the frozen classification head of the optical flow teacher network. The RGB-based DMA representation recovers a large portion of the teacher's motion modeling capacity despite operating without explicit optical flow supervision. This outcome suggests that the RGB backbone can implicitly encode meaningful motion cues through the proposed dynamic filtering mechanism. These findings indicate that effective motion representations can be learned without relying on optical flow inputs.

\vspace{1ex}
\noindent
\textbf{Qualitative Analysis.}
The Grad-CAM visualizations in Fig.~\ref{fig:gradcam_analysis} provide qualitative evidence of the learned motion representations. The optical flow teacher primarily attends to moving regions, while the static Conv3D baseline distributes attention across broader background areas. In contrast, the DMA-enhanced model concentrates on temporally dynamic regions that correspond to action-relevant motion. This observation further confirms that DMA successfully approximates motion-sensitive feature extraction within an RGB-only architecture.

\begin{figure}[htbp]
    \centering
    \begin{subfigure}[b]{0.48\linewidth}
        \centering
        \includegraphics[width=\linewidth]{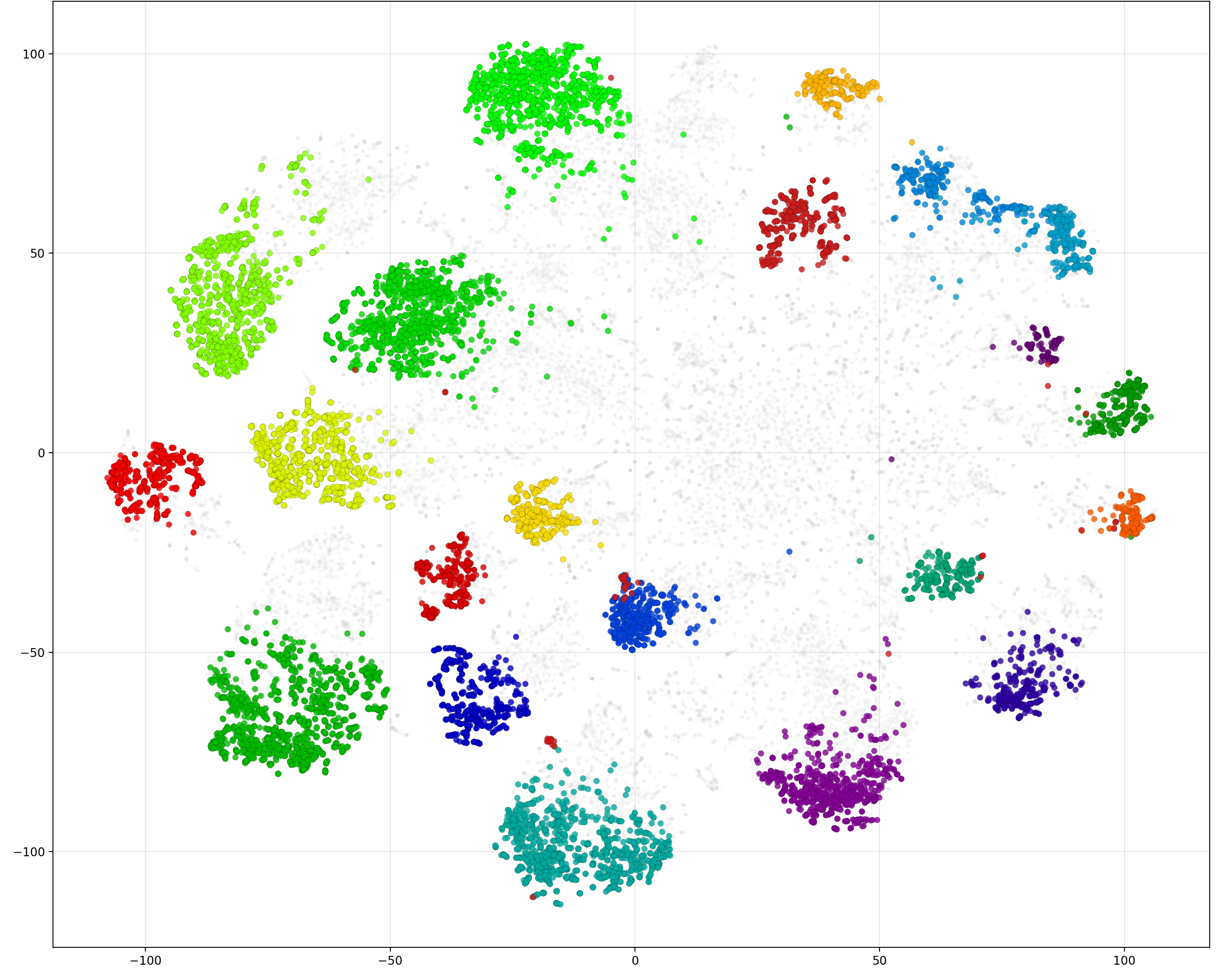}
        \caption{}
        \label{fig:tsne_train}
    \end{subfigure}
    \hfill 
    \begin{subfigure}[b]{0.48\linewidth}
        \centering
        \includegraphics[width=\linewidth]{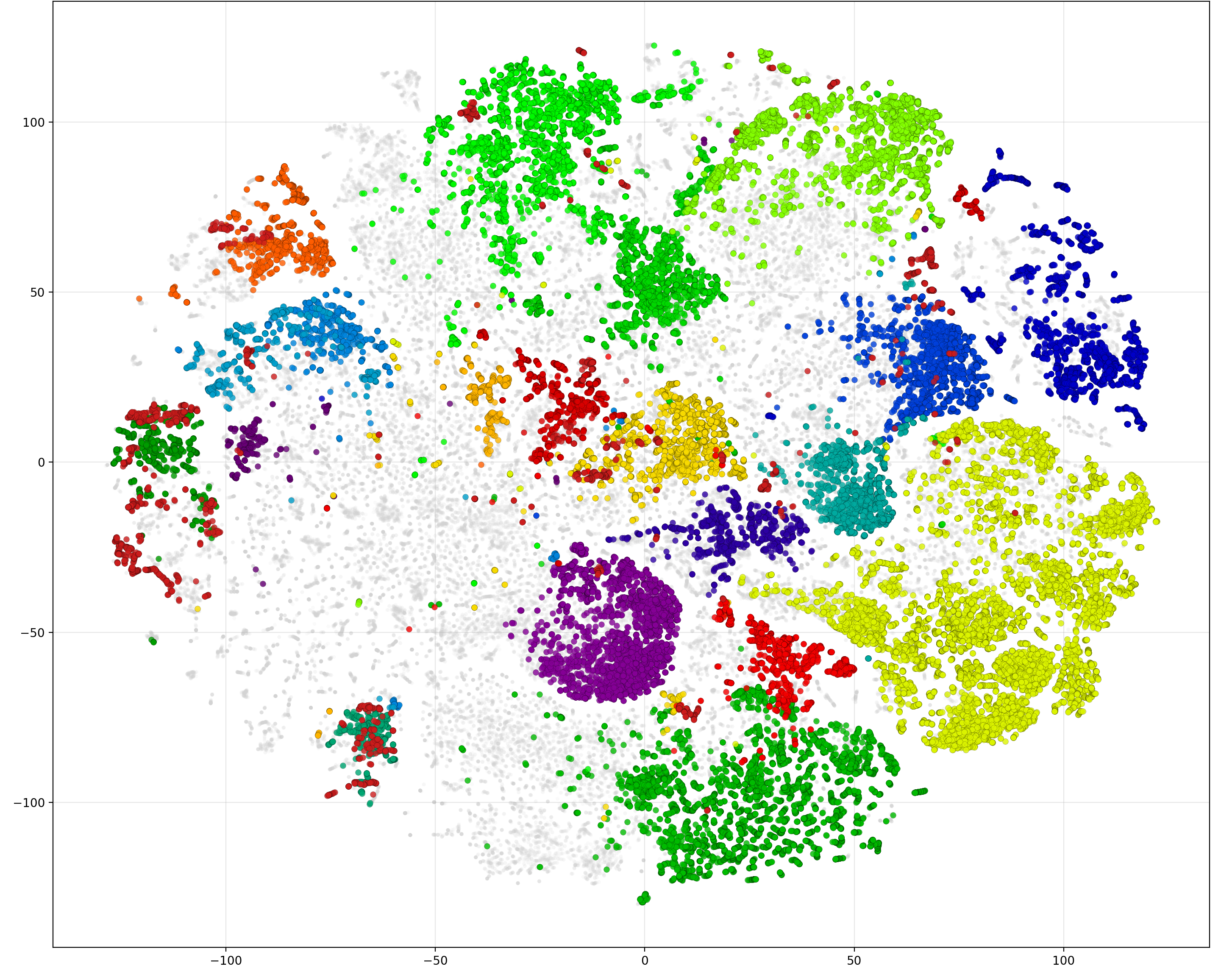}
        \caption{}
        \label{fig:tsne_test}
    \end{subfigure}

    \caption{t-SNE visualization of the learned feature space on THUMOS'14. (a) Training features showing clear semantic clustering produced by Floating SupCon. (b) Test features demonstrating generalization to unseen samples.}
    \label{fig:tsne_vis}
\end{figure}

\vspace{1ex}
\noindent
\textbf{Feature Space Visualization.}
Finally, t-SNE \cite{vandermaaten2008visualizing} visualization in Fig.~\ref{fig:tsne_vis} illustrates the structural effect of Floating SupCon on the learned representation space. The training features form clearly separated semantic clusters corresponding to different actions, while background samples remain more loosely distributed. Importantly, a similar structure appears on the test set, indicating that the learned representation generalizes beyond the training data. These results suggest that Floating SupCon encourages semantically meaningful feature organization while preserving generalization capability.
\section{Discussion}
Despite these promising results, our approach possesses certain limitations. First, the strict reliance on pure RGB inputs makes the system sensitive to adverse environmental conditions, such as poor lighting or sensor noise. Second, the GRU-based temporal module inherently struggles with capturing extremely long-range dependencies compared to pure Transformer architectures. Finally, the momentum-based contrastive learning phase demands substantial memory resources during training due to the large negative sample queue.

Future work will focus on addressing these limitations by exploring linear-time state-space models (e.g., Mamba or S4) to replace the recurrent module for enhanced long-term memory capacity. Additionally, we plan to incorporate domain adaptation techniques to improve robustness against environmental variations, expand the framework to Action Anticipation tasks, and investigate low-bit quantization for direct edge deployment on specialized AI accelerators.

\section{Conclusion}
\label{sec:conclusion}

In this paper, we presented CAKE, a highly efficient real-time framework for Online Action Detection. By introducing the DMA powered by ODConv3D, we successfully distilled complex motion cues into a lightweight RGB backbone, eliminating the need for computationally expensive optical flow during inference. Furthermore, our proposed Floating SupCon effectively resolved the representational mismatch caused by highly diverse background frames. Extensive experiments demonstrated that CAKE achieves state-of-the-art accuracy on the THUMOS'14 and TVSeries datasets while maintaining an impressive real-time inference speed on a single CPU.
\bibliographystyle{IEEEtran}
\bibliography{references}
\end{document}